\documentclass[11pt,a4paper]{article}
\usepackage[dvipdfmx]{graphicx}
\usepackage[nohyperref]{acl2017}
\usepackage{times}
\usepackage{latexsym}
\usepackage{amsfonts}
\usepackage{mathtools}
\usepackage{bm}
\usepackage{lscape}
\usepackage{algorithm,algpseudocode}
\usepackage{url}
\usepackage{xcolor}
\usepackage{algorithm,algpseudocode}

\aclfinalcopy

\title{An Empirical Study of Mini-Batch Creation Strategies\\for Neural Machine Translation}

\author{Makoto Morishita$^{1}$\thanks{\ \ \ This work is done while the author was at Nara Institute of Science and Technology.}, \ 
  Yusuke Oda$^{2}$, 
  Graham Neubig$^{3,2}$,\\
  \textbf{Koichiro Yoshino$^{2,4}$, 
  Katsuhito Sudoh$^{2}$, 
  Satoshi Nakamura$^{2}$}\\
  $^{1}$NTT Communication Science Laboratories, NTT Corporation\\
  $^{2}$Nara Institute of Science and Technology\\
  $^{3}$Carnegie Mellon University\\
  $^{4}$PRESTO, Japan Science and Technology Agency\\
  {\tt morishita.makoto@lab.ntt.co.jp},\ \ {\tt gneubig@cs.cmu.edu} \\
  {\tt \{oda.yusuke.on9, koichiro, sudoh, s-nakamura\}@is.naist.jp} \\
  }

\date{}

\begin{document}
\maketitle
\begin{abstract}
Training of neural machine translation (NMT) models
usually uses mini-batches for efficiency purposes.
During the mini-batched training process, it is necessary to pad shorter sentences in a mini-batch to be equal in length to the longest sentence therein for efficient computation.
Previous work has noted that sorting the corpus based on the sentence length before making mini-batches reduces the amount of padding and increases the processing speed.
However, despite the fact that mini-batch creation is an essential step in NMT training, widely used NMT toolkits implement disparate strategies for doing so,
which have not been empirically validated or compared.
This work investigates mini-batch creation strategies with experiments over two different datasets.
Our results suggest that the choice of a mini-batch creation strategy has a large effect on NMT training and some length-based sorting strategies do not always work well compared with simple shuffling.
\end{abstract}

\section{Introduction}
\label{sec:introduction}

Mini-batch training is a standard practice in large-scale machine learning.  In recent implementations of neural networks, the efficiency of loss and gradient calculation is greatly improved by mini-batching due to the fact that combining training examples into batches allows for fewer but larger operations that can take advantage of the parallelism allowed by modern computation architectures, particularly GPUs.

In some cases, such as the case of processing images, mini-batching is straightforward, as the inputs in all training examples take the same form. However, in order to perform mini-batching in the training of neural machine translation (NMT) or other sequence-to-sequence models,

we need to {\em pad} shorter sentences to be the same length as the longest sentences
to account for sentences of variable length in each mini-batch.

To help prevent wasted calculation due to this padding, it is common to sort the corpus according to the sentence length before creating mini-batches \cite{sutskever14sequencetosequence,bahdanau15alignandtranslate}, because putting sentences that have similar lengths in the same mini-batch will reduce the amount of padding and increase the per-word computation speed.
However, we can also easily imagine that this grouping of sentences together may affect the convergence speed and stability, and the performance of the learned models.
Despite this fact, no previous work has explicitly examined how mini-batch creation affects the learning of NMT models. Various NMT toolkits include implementations of different strategies, but they have neither been empirically validated nor compared.

In this work, we attempt to fill this gap by surveying the various mini-batch creation strategies that are in use: sorting by length of the source sentence, target sentence, or both, as well as making mini-batches according to the number of sentences and the number of words.
We empirically compare their efficacy on two translation tasks and find that some strategies in wide use are not necessarily optimal for reliably training models.

\section{Mini-batches for NMT}
\label{sec:mini_batch_for_nmt}

\begin{figure*}[!tb]
  \centering
  \includegraphics[width=0.95\linewidth]{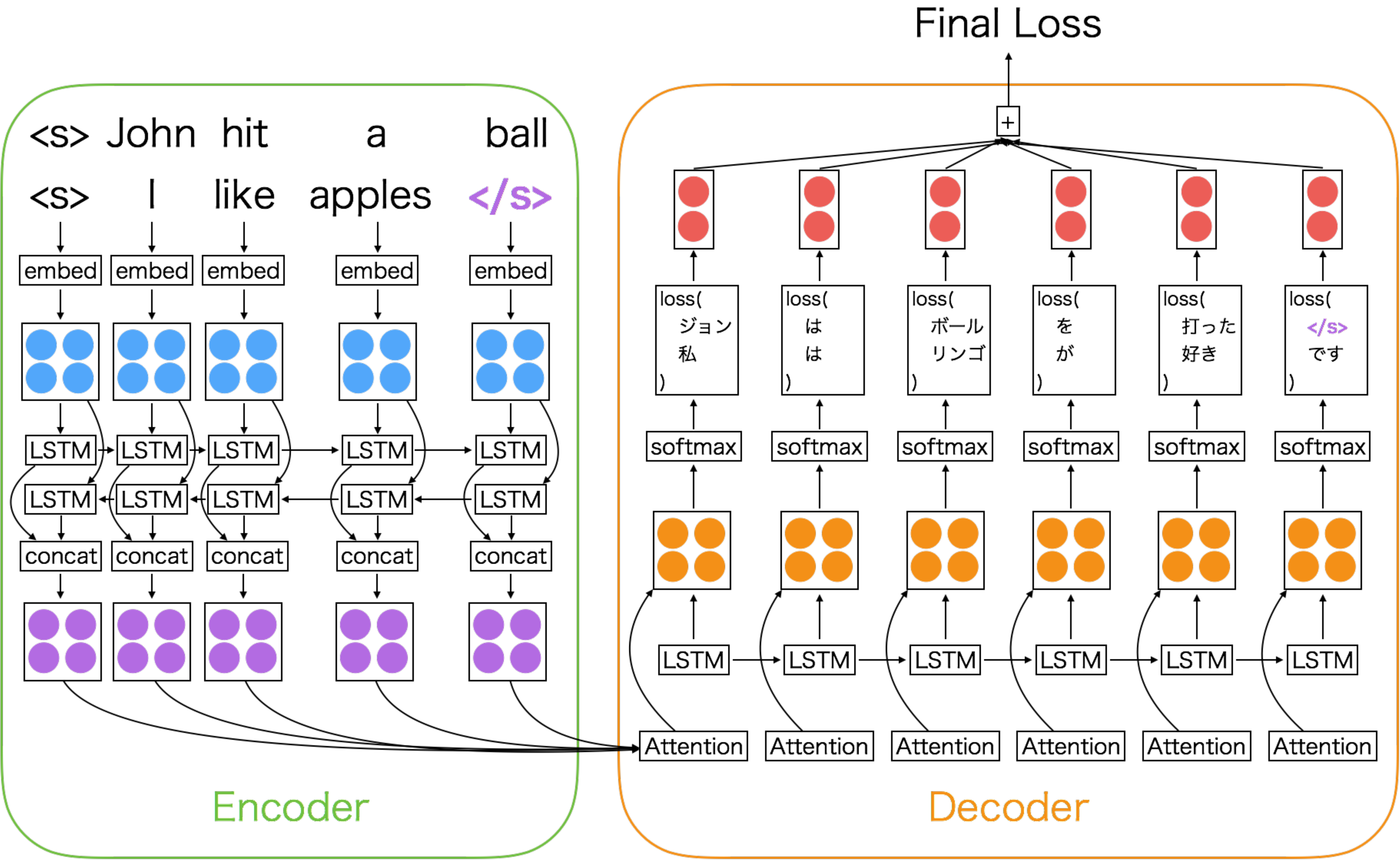}
  \caption{An example of mini-batching in an encoder-decoder translation model.}
  \label{fig:mini-batch_enc-dec}
\end{figure*}

First, to clearly demonstrate the problem of mini-batching in NMT models, Figure~\ref{fig:mini-batch_enc-dec} shows an example of mini-batching two sentences of different lengths in an encoder-decoder model.

The first thing that we can notice from the figure is that multiple operations at a particular time step $t$ can be combined into a single operation.
For example, both ``John'' and ''I'' are embedded in a single step into a matrix that is passed into the encoder LSTM in a single step.
On the target side as well, we calcualate the loss for the target words at time step $t$ for every sentence in the mini-batch simultaneously. 

However, there are problems when sentences are of different length, as only some sentences will have any content at a particular time step.
To resolve this problem, we pad short sentences with end-of-sentence tokens to adjust their length to the length of the longest sentence.
In the Figure~\ref{fig:mini-batch_enc-dec}, purple colored ``$\langle$/s$\rangle$'' indicates the padded end-of-sentence token.

Padding with these tokens makes it possible to handle variably-lengthed sentences as if they were of the same length.
On the other hand, the computational cost for a mini-batch increases in proportion to the longest sentence therein, and excess padding can result in a significant amount of wasted computation.
One way to fix this problem is by creating mini-batches that include sentences of similar length \cite{sutskever14sequencetosequence} to reduce the amount of padding required.
Many NMT toolkits implement length-based sorting of the training corpus for this purpose.
In the following section, we discuss several different mini-batch creation strategies used in existing neural MT toolkits.

\section{Mini-batch Creation Strategies}
\label{sec:mini_batch_creation_strategies}
\begin{algorithm}[tb]
\caption{Create mini-batches}\label{alg:minibatch}
\begin{algorithmic}[1]
\State $\bm{C}\gets {\rm Training\ corpus}$
\State $\bm{C}\gets$ sort$(\bm{C})$ or shuffle$(\bm{C})$\Comment{sort or shuffle the whole corpus}
\State $\bm{B}\gets \{\}$\Comment{mini-batches}
\State $i\gets 0,\ j\gets 0$
\While{$i<\bm{C}.$size()}
\State $\bm{B}[j]\gets \bm{B}[j] + \bm{C}[i]$
\If{$\bm{B}[j].$size() $\ge$ max mini-batch size}
\State $\bm{B}[j]\gets$ padding$(\bm{B}[j])$\Comment{Padding tokens to the longest sentence in the mini-batch}
\State $j\gets j + 1$
\EndIf
\State $i\gets i + 1$
\EndWhile
\State $\bm{B}\gets$ shuffle$(\bm{B})$\Comment{shuffle the order of the mini-batches}
\end{algorithmic}
\end{algorithm}

Specifically, we examine three aspects of mini-batch creation: mini-batch size, word vs. sentence mini-batches, and sorting strategies.
Algorithm \ref{alg:minibatch} shows the pseudo code of creating mini-batches.

\subsection{Mini-batch Size}
\label{sec:mini_batch_size}
The first aspect we consider is \textit{mini-batch size} for which, of the three aspects we examine here, the effect is relatively well known.

When we use larger mini-batches, more sentences participate in the gradient calculation making the gradients more stable.
They also increase efficiency with parallel computation.
However, they decrease the number of parameter updates performed in a certain amount of time, which can slow convergence at the beginning of training.
Large mini-batches can also pose problems in practice due to the fact that they increase memory requirements.

\subsection{Sentence vs. Word Mini-batching}
\label{sec:mini_batch_unit}
The second aspect that we examine, which has not been examined in detail previously, is whether to create mini-batches based on the \textit{number of sentences} or \textit{number of target words}.

Most NMT toolkits create mini-batches with a constant number of sentences.
In this case, the number of words included in each mini-batch differs greatly due to the variance in sentence lengths.
If we use the neural network library that constructs graphs in a dynamic fashion (e.g. DyNet \cite{neubig17dynet}, Chainer \cite{tokui15chainer}, or PyTorch\footnote{\url{http://pytorch.org}}), this will lead to a large variance in memory consumption from mini-batch to mini-batch.
In addition, because the loss function for the mini-batch is equal to the sum of the losses incurred for each word, the scale of the losses will vary greatly from mini-batch to mini-batch, which could be potentially detrimental to training.

Another choice is to create mini-batches by keeping the number of target words in each mini-batch approximately stable, but varying the number of sentences. 
We hypothesize that this may lead to more stable convergence, and test this hypothesis in the experiments.

\subsection{Corpus Sorting Methods}
\label{sec:sorting_methods}

The final aspect that we examine, which has similarly is not yet well understood, is the effect of \textit{the method that we use to sort the corpus before grouping consecutive sentences into mini-batches}.

A standard practice in online learning shuffles training samples
to ensure that bias in the presentation order does not adversely affect the final result.
However, 
as we mentioned in Section~\ref{sec:mini_batch_for_nmt}, NMT studies \cite{sutskever14sequencetosequence,bahdanau15alignandtranslate} prefer uniform length samples in the mini-batch by sorting the training corpus, to reduce the amount of padding and increase per-word calculation speed.
In particular, in the encoder-decoder NMT framework \cite{sutskever14sequencetosequence},
the computational cost in the softmax layer of the decoder is much heavier than the encoder.
Some NMT toolkits sort the training corpus based on the target sentence length to avoid unnecessary softmax computations on padded tokens in the target side.
Another problem arises in the attentional NMT model \cite{bahdanau15alignandtranslate,luong15emnlp};
attentions may give incorrect positive weights to the padded tokens in the source side.
The problems above also motivate the mini-batch creation with uniform length sentences with fewer padded tokens.

Inspired by sorting methods in use in current open source implementations, we compare the following sorting methods:

\begin{description}
  \setlength{\parskip}{0cm}
  \setlength{\itemsep}{0cm}
  \item[\textsc{shuffle}:] Shuffle the corpus randomly before creating mini-batches, with no sorting.
  \item[\textsc{src}:] Sort based on the source sentence length. 
  \item[\textsc{trg}:] Sort based on the target sentence length. 
  \item[\textsc{src\_trg}:] Sort using the source sentence length, break ties by sorting by target sentence length.
  \item[\textsc{trg\_src}:] Converse of \textsc{src\_trg}.
\end{description}

Of established open-source toolkits,
OpenNMT \cite{klein17opennmt} uses the \textsc{src} sorting method,
Nematus\footnote{\url{https: //github.com/rsennrich/nematus}} and KNMT \cite{cromieres16knmt} use the \textsc{trg} sorting method,
and lamtram\footnote{\url{https: //github.com/neubig/lamtram}} uses the \textsc{trg\_src} sorting method.

\section{Experiments}
\label{sec:experiments}
\begin{table}[tb] 
\centering
\hbox to\hsize{\hfil
\begin{tabular}{c|r|r}
 & ASPEC-JE & WMT 2016 \\ \hline
train & 2,000,000 & 4,562,102 \\
dev & 1,790 & 2,169 \\
test & 1,812 & 2,999 \\
\end{tabular}\hfil}
\caption{Number of sentences in the corpus} \label{tab:corpus_size}
\end{table}

We conducted NMT experiments with the strategies presented above to examine their effects on NMT training.
\subsection{Experimental Settings}
\label{sec:experimental_settings}
We carried out experiments with two language pairs, English-Japanese using the ASPEC-JE corpus \cite{nakazawa16aspec} and English-German using the WMT 2016 news task with news-test2016 as the test-set \cite{bojar16wmt}.
Table \ref{tab:corpus_size} shows the number of sentences contained in the corpora.

The English and German texts were tokenized with {\tt tokenizer.perl}\footnote{\url{https://github.com/moses-smt/mosesdecoder/blob/master/scripts/tokenizer/tokenizer.perl}}, and the Japanese texts were tokenized with KyTea \cite{neubig11aclshort}.

As a testbed for our experiments, we used the standard global attention model of \newcite{luong15emnlp} with attention feeding and a bidirectional encoder with one LSTM layer of 512 nodes.
We used the DyNet-based \cite{neubig17dynet} NMTKit\footnote{\url{https://github.com/odashi/nmtkit} We used the commit number {\tt 566e9c2}.}, with a vocabulary size of 65536 words and dropout of 30\% for all vertical connections.
We used the same random numbers as initial parameters for each experiment to reduce variance due to initialization.
We used Adam \cite{kingma14adam} ($\alpha=0.001$) or SGD ($\eta=0.1$) as the learning algorithm.
After every 50,000 training sentences, we processed the test set to record negative log likelihoods.
In the testing, we set the mini-batch size to 1, in order to calculate negative log likelihood correctly.
We calculated the case-insensitive BLEU score \cite{papineni02bleu} with {\tt multi-bleu.perl}\footnote{\url{https://github.com/moses-smt/mosesdecoder/blob/master/scripts/generic/multi-bleu.perl}} script.

Table \ref{tab:comparative_approach} shows the mini-batch creation settings compared in this paper,
and we tried all sorting methods discussed in Section~\ref{sec:sorting_methods} for each setting.
In method (e), we set the average number of target words in 64 sentences: 2055 words for ASPEC-JE, 1742 words for WMT.
For all experiments, we shuffled the processing order of the mini-batches.

\begin{table}[tb] 
\centering
\hbox to\hsize{\hfil
\begin{tabular}{c|rr}
 & mini-batch units & learning algorithm \\ \hline
(a) & 64 sentences & Adam \\
(b) & 32 sentences & Adam \\
(c) & 16 sentences & Adam \\
(d) & 8 sentences & Adam \\
(e) & 2055 or 1742 words & Adam \\
(f) & 64 sentences & SGD \\
\end{tabular}\hfil}
\caption{Compared settings} \label{tab:comparative_approach}
\end{table}

\subsection{Experimental Results and Analysis}
\label{sec:experimental_results_and_analysis}
\begin{figure*}[!tb]
  \centering
  \includegraphics[width=0.95\linewidth]{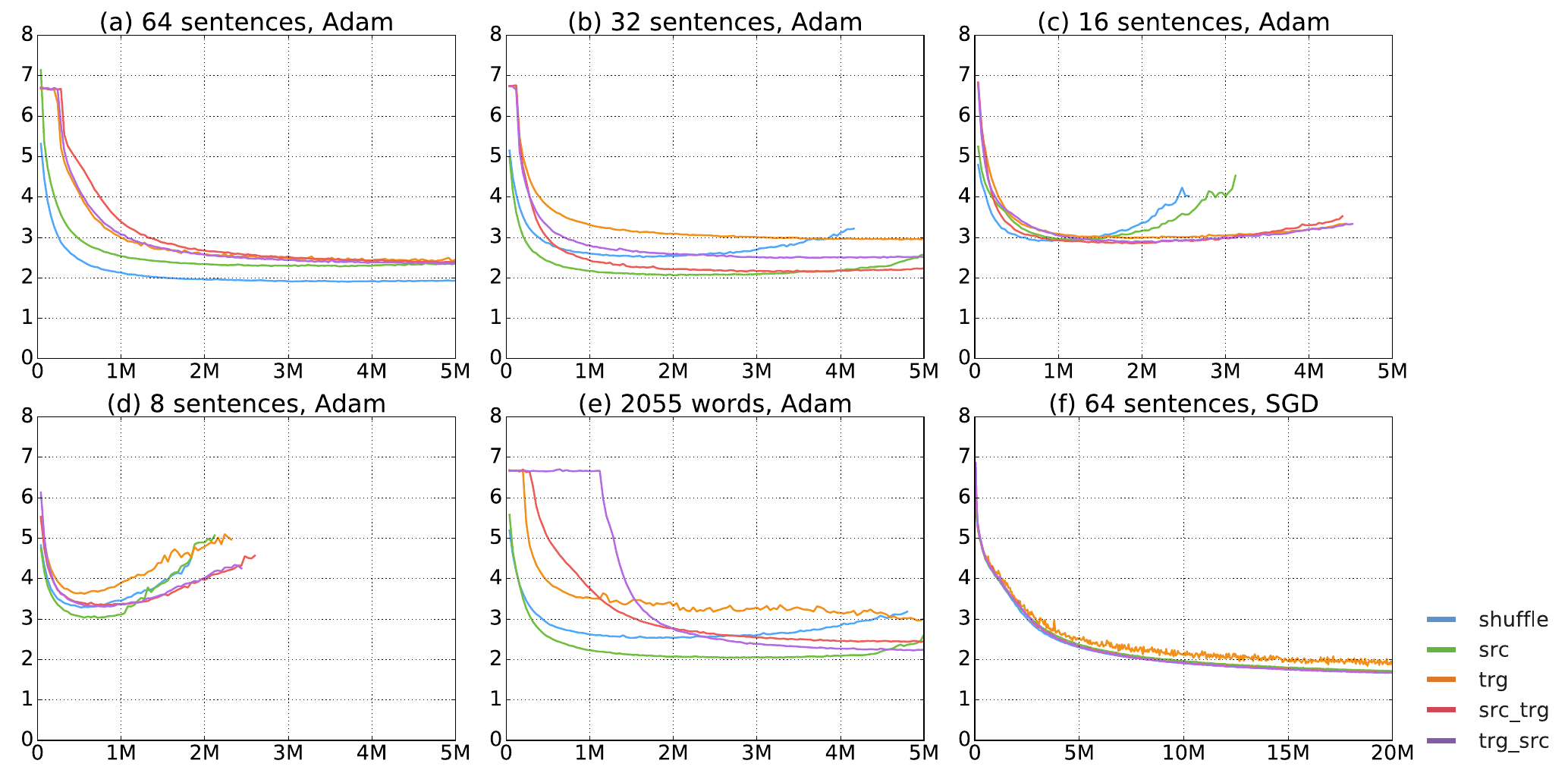}
  \caption{Training curves on the ASPEC-JE test set. The y- and x-axes shows the negative log likelihoods and number of processed sentences. The scale of the x-axis in the method (f) is different from others.}
  \label{fig:aspec_test_ppl_graphs}
\end{figure*}

\begin{figure*}[!tb]
  \centering
  \includegraphics[width=0.95\linewidth]{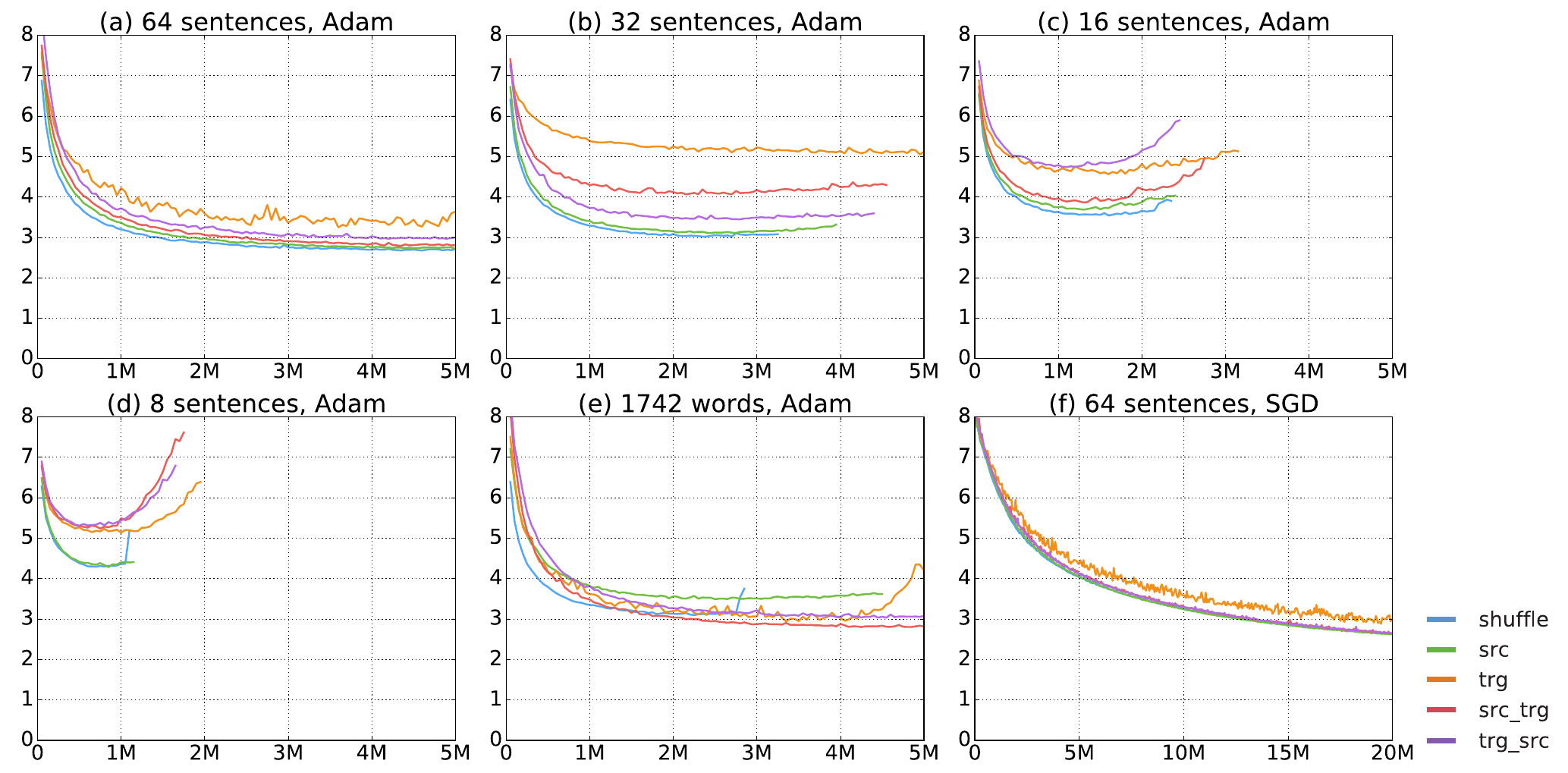}
  \caption{Training curves on the WMT2016 test set. Axes are the same as Figure~\ref{fig:aspec_test_ppl_graphs}.}
  \label{fig:wmt_test_ppl_graphs}
\end{figure*}

\begin{figure*}[!tb]
  \centering
  \includegraphics[width=0.95\linewidth]{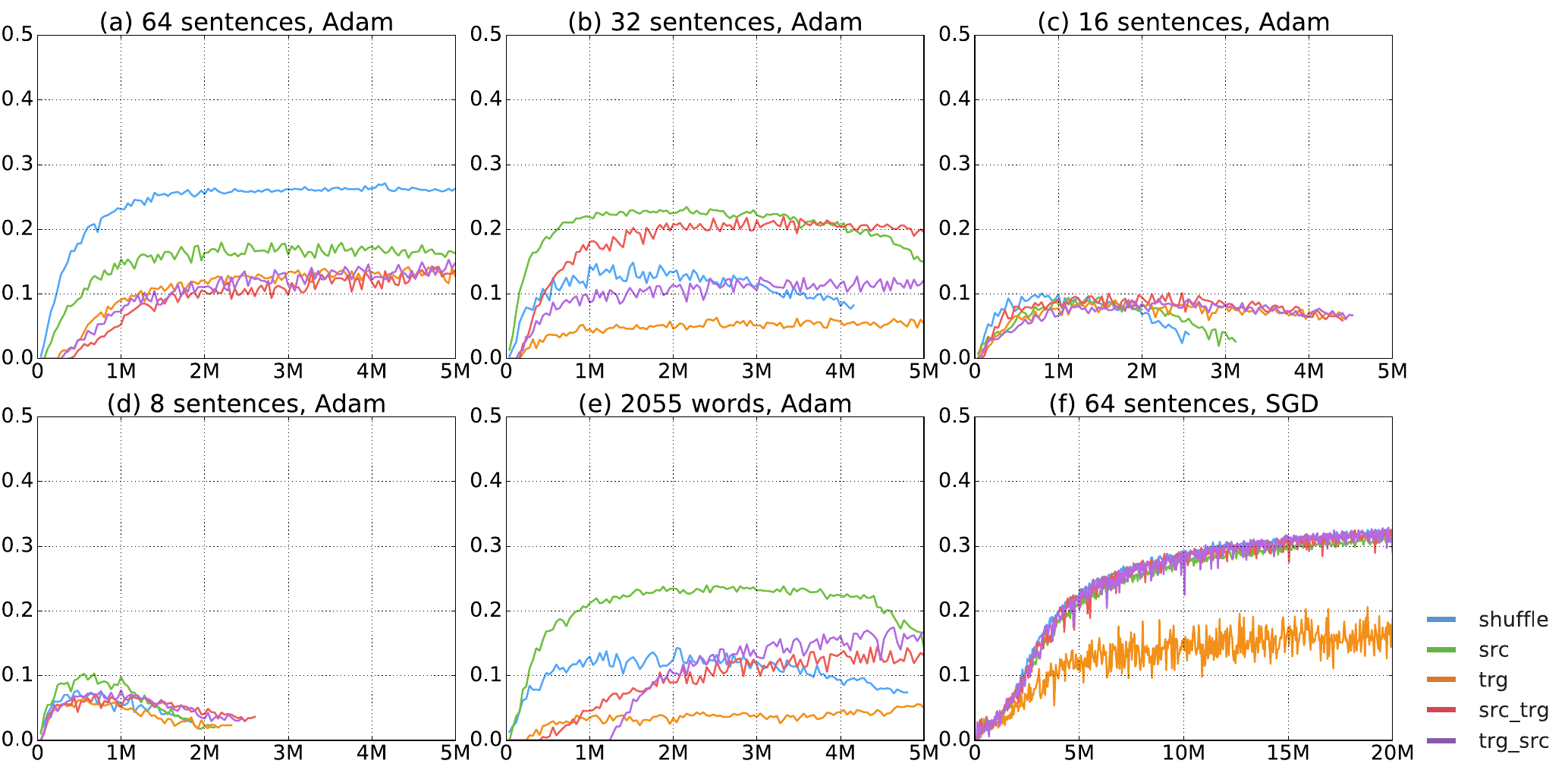}
  \caption{BLEU scores on the ASPEC-JE test set. The y- and x-axes shows the BLEU scores and number of processed sentences. The scale of the x-axis in the method (f) is different from others.}
  \label{fig:aspec_test_bleu_graphs}
\end{figure*}

\begin{figure*}[!tb]
  \centering
  \includegraphics[width=0.95\linewidth]{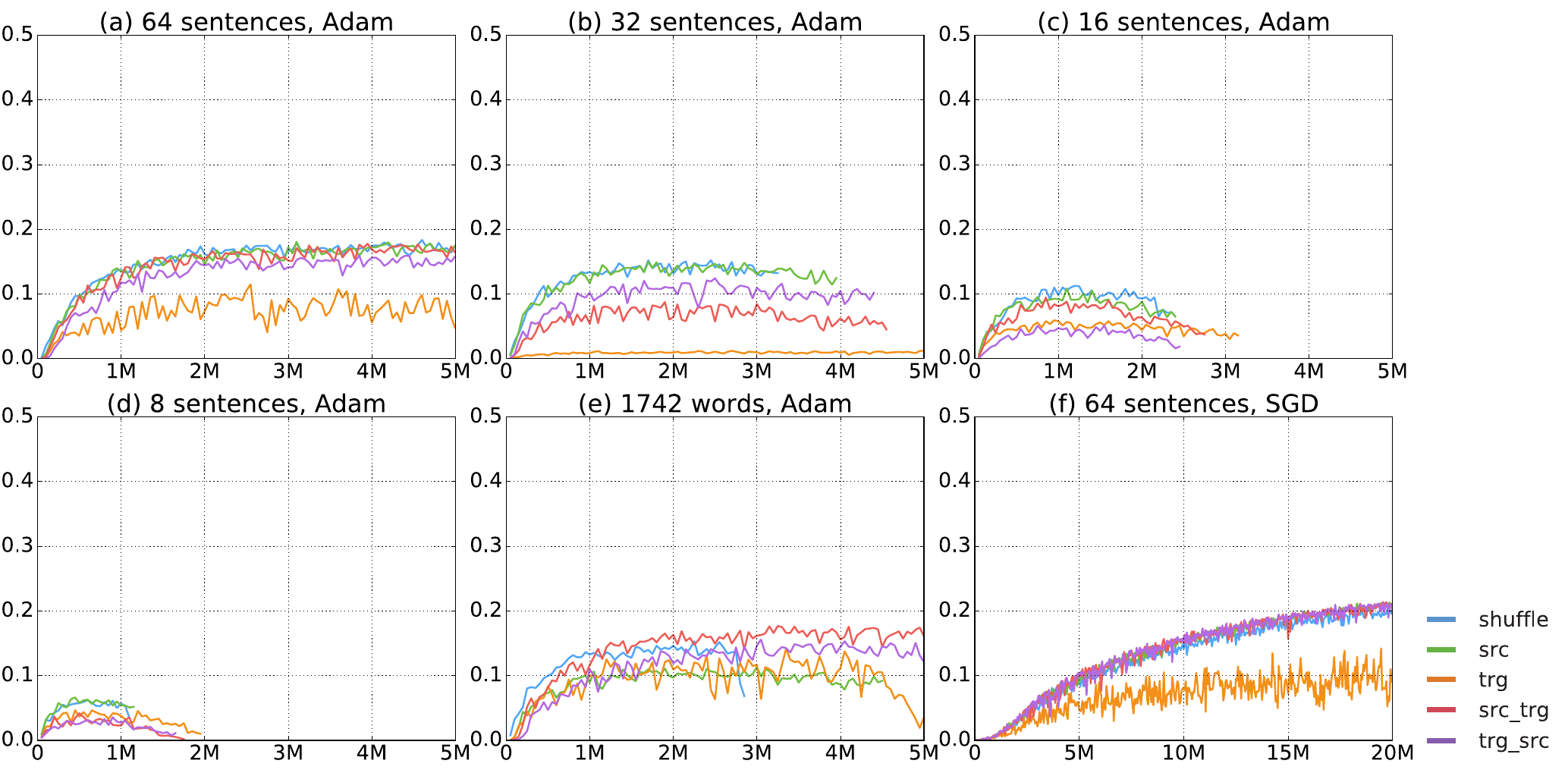}
  \caption{BLEU scores on the WMT2016 test set. Axes are the same as Figure~\ref{fig:aspec_test_bleu_graphs}.}
  \label{fig:wmt_test_bleu_graphs}
\end{figure*}

Figure~\ref{fig:aspec_test_ppl_graphs}, \ref{fig:wmt_test_ppl_graphs}, \ref{fig:aspec_test_bleu_graphs} and \ref{fig:wmt_test_bleu_graphs} show the transition of negative log likelihoods and the BLEU scores according to the number of processed sentences in ASPEC-JE and WMT2016 test sets.
Table \ref{tab:corpus_time} shows the average time to process the whole ASPEC-JE corpus.

The learning curves show very similar tendencies in different language pairs.
We discuss the results in detail on each strategy that we investigated.



\begin{table}[tb] 
\centering
\hbox to\hsize{\hfil
\begin{tabular}{cc}
sorting method & average time (hour) \\ \hline
\textsc{shuffle} & 8.08 \\
\textsc{src} & 6.45 \\
\textsc{trg} & 5.21 \\
\textsc{src\_trg} & 4.35 \\
\textsc{trg\_src} & 4.30 \\
\end{tabular}\hfil}
\caption{Average time needed to train a whole ASPEC-JE corpus using method (a). We used a GTX 1080 GPU for this experiment.} \label{tab:corpus_time}
\end{table}

\subsubsection{Effect of Mini-batch Size}
\label{sec:effect_size}
We carried out the experiments with the mini-batch size of 8 to 64 sentences.\footnote{We tried the experiments with larger mini-batch size, but we couldn't run it due to the GPU memory limitation.}

From the experimental results of the method (a), (b), (c) and (d), in the case of using Adam, the mini-batch size affects the training speed and it also has an impact on the final accuracy of the model.
As we mentioned in Section~\ref{sec:mini_batch_size}, the gradients can be stabler by increasing the mini-batch size, and it seems to have a positive impact on the model from the view of accuracy.
Thus, we can first note that mini-batch size is a very important hyper-parameter for NMT training that should not be ignored.
In our case in particular, the largest mini-batch size that could be loaded into the memory was the best for the NMT training.

\subsubsection{Effect of Mini-batch Unit}
\label{sec:effect_unit}
Looking at the experimental results of the methods (a) and (e),  we can see that perplexities drop faster if we use \textsc{shuffle} for method (a) and \textsc{src} for method (e), but we couldn't see any large differences in terms of the training speed and the final accuracy of the model.
We hypothesize that the large variance of the loss affects the final model accuracy, especially when using the learning algorithm that uses momentum such as Adam.
However, these results indicate that these differences do not significantly affect the training results.
We leave a comparison of memory consumption for future research.

\subsubsection{Effect of Corpus Sorting Method using Adam}
\label{sec:effect_sort_adam}
From all experimental results of the method (a), (b), (c), (d) and (e), in the case of using \textsc{shuffle} or \textsc{src}, perplexities drop faster and tend to converge to lower perplexities than the other methods for all mini-batch sizes.
We believe the main reason for this is due to the similarity of the sentences contained in each mini-batch.
If the sentence length is similar, the features of the sentence may also be similar.
We carefully examined the corpus and found that at least this is true for the corpus we used (e.g. shorter sentences tend to contain the similar words).
In this case, if we sort sentences by their length, sentences that have similar features will be gathered into the same mini-batch, making training less stable than if all sentences in the mini-batch had different features.
This is evidenced by the more jagged lines of the \textsc{trg} method.

As a conclusion, the \textsc{trg} and \textsc{trg}\_\textsc{src} sorting methods, which are used by many NMT toolkits, have a higher overall throughput when just measuring the number of words processed,
but for convergence speed and final model accuracy, it seems to be better to use \textsc{shuffle} or \textsc{src}.

Some toolkits shuffle the corpus first, then create mini-batches by sorting a few consecutive sentences.
We think that this method may be effective by combining the advantage of \textsc{shuffle} and other sorting methods, but an empirical comparison is beyond the scope of this work.

\subsubsection{Effect of Corpus Sorting Method using SGD}
\label{sec:effect_sort_sgd}
By comparing the experimental results of the methods (a) and (f),  we found that in the case of using Adam, the learning curves greatly depend on the sorting method, but in the case of using SGD there was little effect.
This is likely because SGD makes less bold updates of rare parameters, improving its overall stability.
However, we find that only when using the \textsc{trg} method, the negative log likelihoods and the BLEU scores are not stable.
It can be conjectured that this is an effect of gathering the similar sentences in a mini-batch as we mentioned in Section~\ref{sec:effect_sort_adam}.
These results indicate that in the case of SGD it is acceptable to \textsc{trg}\_\textsc{src}, which is the fastest method to process the whole corpus (see Table \ref{tab:corpus_time}), for SGD.

Recently, \newcite{yonghui16gnmt} proposed a new learning paradigm, which uses Adam for the initial training, then switches to SGD after several iterations.
If we use this learning algorithm, we may be able to train the model more effectively by using \textsc{shuffle} or \textsc{src} sorting method for Adam, and \textsc{trg}\_\textsc{src} for SGD.

\subsection{Experiments with a Different Toolkit}
\begin{figure}[!tb]
  \centering
  \includegraphics[width=1.00\linewidth]{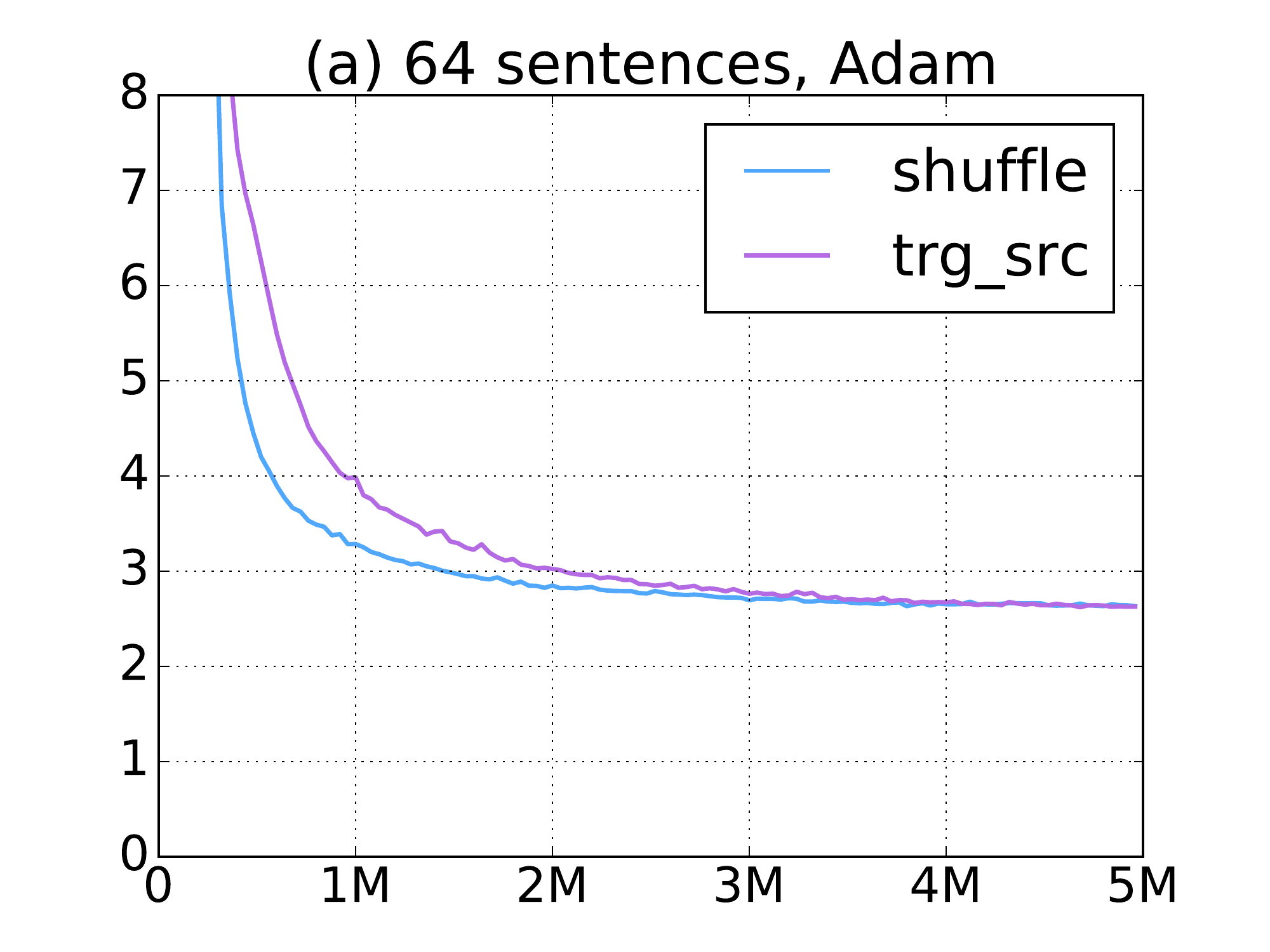}
  \caption{Training curves on the ASPEC test set using lamtram toolkit. Axes are the same as Figure~\ref{fig:aspec_test_ppl_graphs}.}
  \label{fig:lamtram_ppl_graph}
\end{figure}
In the previous experiments, we conducted the experiments with only one NMT toolkit, so the results may be dependent on the particular implementation provided therein.
To ensure that these results generalize to other toolkits with different default parameters, we conducted the experiments with another NMT toolkit.

\subsubsection{Experimental Settings}
In this section, we used lamtram\footnote{\url{https: //github.com/neubig/lamtram}} as a NMT toolkit.
We carried out the Japanese-English translation experiments with ASPEC-JE corpus.
We used Adam \cite{kingma14adam} ($\alpha=0.001$) as the learning algorithm and tried the two sorting algorithms: \textsc{shuffle} which is the best sorting method on previous experiments and \textsc{trg\_src} which is the default sorting method used by the lamtram toolkit. 
Normally, lamtram creates mini-batches based on the number of target words contained in each mini-batch, but we changed it to fix the mini-batch size to 64 sentences because we find that larger mini-batch size seems to be better in the previous experiments.
Other experimental settings are the same as described in the Section~\ref{sec:experimental_settings}.

\subsubsection{Experimental Results}
Figure~\ref{fig:lamtram_ppl_graph} shows the transition of negative log likelihoods using lamtram.
We can see the tendency of the training curves are similar to the Figure~\ref{fig:aspec_test_ppl_graphs} (a), the combination with \textsc{shuffle} drops negative log likelihood faster than the \textsc{trg\_src} one.

From this experiments, we could verify that our experimental results in the Section~\ref{sec:experiments} do not rely on the toolkit and we think the observed behavior will generalize to other toolkits and implementations.

\section{Related Work}
Recently, \newcite{britz17massive} have released a paper about exploring the hyper-parameters of NMT.
This work is similar to our paper in the terms of finding the better hyper-parameters by doing a large number of experiments and deriving empirical conclusions.
However, notably this paper fixed the mini-batch size to 128 sentences and did not treat mini-batch creation strategy as one of the hyper-parameters of the model.
With our experimental results, we argue that the mini-batch creation strategies also have an impact on the NMT training,
and thus having solid recommendations for how to adjust this hyper-parameter are also of merit.

\section{Conclusion}
\label{sec:conclusion}
In this paper, we analyzed how mini-batch creation strategies affect the training of NMT models for two language pairs.
The experimental results suggest mini-batch creation strategy is an important hyper-parameter of the training process,
and commonly-used sorting strategies are not always optimal.
We sum up the results as follows:
\begin{itemize}
	\item Mini-batch size can affect the final accuracy of the model in addition to the training speed and the larger mini-batch size seems to be better.
    \item Mini-batch units do not effect to the training process, so it is possible to use either the number of sentences or target words.
    \item We should use \textsc{shuffle} or \textsc{src} sorting method for Adam, and it is sufficient to use \textsc{trg}\_\textsc{src} for SGD.
\end{itemize}

In the future, we plan to do experiments with larger mini-batch sizes and compare the used peak memory between making mini-batches by the number of sentences or target words.
We are also interested in checking the effects of different mini-batch creation strategies with other language pairs, corpora and optimization functions.

\section*{Acknowledgments}
This work was done as a part of the joint research project with NTT and Nara Institute of Science and Technology.
This research has been supported in part by JSPS KAKENHI Grant Number 16H05873.
We thank the anonymous reviewers for their insightful comments.

\bibliography{myplain,main}
\bibliographystyle{acl_natbib}

\end{document}